\documentclass[letterpaper, 10 pt, conference]{ieeeconf}  
\IEEEoverridecommandlockouts                              
\usepackage{graphics}    
\usepackage{times}       
\usepackage{amsmath}     
\usepackage{amssymb}     
\usepackage{graphicx}
\usepackage{algorithm}
\usepackage[noend]{algpseudocode}
\usepackage{xspace}
\usepackage{booktabs}
\usepackage{multirow}

\def\secref#1{Sec.~\ref{#1}}
\def\figref#1{Fig.~\ref{#1}}
\def\tabref#1{Tab.~\ref{#1}}
\def\eqref#1{Eq.~(\ref{#1})}

\newcommand\etal{\emph{et al.}\xspace}

\def\argmax{\mathop{\rm argmax}}

\newcommand{\cropvsweed}{crop-weed\xspace}
\newcommand{\Cropvsweed}{Crop-weed\xspace}

\renewcommand{\vec}[1]{\mathbf{#1}}

\renewcommand{\vec}[1]{\mathbf{#1}}	
\newcommand{\set}[1]{\mathcal{#1}} 	

\newcommand{\dataA}{BoniRob~dataset\xspace} 
\newcommand{\dataB}{UAV~dataset\xspace}

\newcommand{\grass}{grass weed\xspace}
\newcommand{\grasses}{grass weeds\xspace}

\newcommand{\dicotyl}{dicot weed\xspace}
\newcommand{\dicotyls}{dicot weeds\xspace}

\usepackage{array}
\newcolumntype{L}[1]{>{\raggedright\let\newline\\\arraybackslash\hspace{0pt}}m{#1}}
\newcolumntype{C}[1]{>{\centering\let\newline\\\arraybackslash\hspace{0pt}}m{#1}}
\newcolumntype{R}[1]{>{\raggedleft\let\newline\\\arraybackslash\hspace{0pt}}m{#1}}

\usepackage{fancyhdr}
\title{\LARGE \bf Joint Stem Detection and Crop-Weed Classification\\ for Plant-specific Treatment in Precision Farming}

\author{Philipp Lottes \and Jens Behley  \and Nived Chebrolu \and Andres Milioto \and Cyrill Stachniss
  \thanks{All authors are with the University of Bonn, Germany. }%
  \thanks{This work has partly been supported by the EC under the grant number H2020-ICT-644227-Flourish. 
  }%
}

\begin{document}
\maketitle
\thispagestyle{fancy}
\pagestyle{fancy}

\begin{abstract}
 
  Applying agrochemicals is the default procedure for conventional weed control in 
  crop production, but has negative impacts on the environment. Robots
  have the potential to treat every plant in the field individually and thus
  can reduce the required use of such chemicals. To achieve that, robots need 
  the ability to identify crops and weeds in the field and must additionally 
  select effective treatments. While certain types of weed can be 
  treated mechanically, other types need to be treated by (selective) 
  spraying. 
  In this paper, we present an approach that provides the necessary information for 
  effective plant-specific treatment. It outputs the stem location for weeds, which allows  
  for mechanical treatments, and the covered area of the weed for selective spraying.
  Our approach uses an end-to-end trainable fully convolutional network that 
  simultaneously estimates stem positions as well as the covered area of crops
  and weeds.  It jointly learns the class-wise stem detection and the pixel-wise 
  semantic segmentation.
  Experimental evaluations on different real-world datasets show that our
  approach is able to reliably solve this problem. Compared to state-of-the-art approaches, our approach not only substantially improves the stem
  detection accuracy, i.e., distinguishing crop and weed stems, but also
  provides an improvement in the semantic segmentation performance.
\end{abstract}

\section{Introduction}
\label{sec:intro}

Agrochemicals such as pesticides, herbicides, and fertilizer are currently
needed in conventional agriculture for effective weed control and attaining
high yields. Agrochemicals, however, can have a negative impact on the
environment and consequently affect human health. Thus, sustainable
crop production should reduce the amount of applied chemicals. Today, weed
control avoiding  agrochemicals, however, is often a manual and 
labor-intensive task.

Robots for precision farming offer a great potential to address this challenge
through a targeted treatment on a per-plant level. Agricultural robots equipped
with different actuators for weed control such as selective sprayers,
mechanical tools, or even lasers, can execute the treatments only where it is
actually needed and also can select the most effective treatment for the
targeted  plant or weed. For example, mechanical and laser-based treatments
are most effective if applied to the stem location of a plant. In contrast,
grass-like weeds are most effectively treated by applying agrochemicals
on their entire leaf area.

To realize a selective and plant-dependent treatment, farming robots need an
effective plant classification system. Such a system needs to reliably
identify both, the stem location of \dicotyls (weeds whose seeds having
two embryonic leaves) and also the extent of grass weeds given by
its leaf area. In
this paper, we thus address exactly this problem so that a 
robot can perform targeted, plant-specific
treatments.

\begin{figure}
  \centering
 \includegraphics[width=\linewidth]{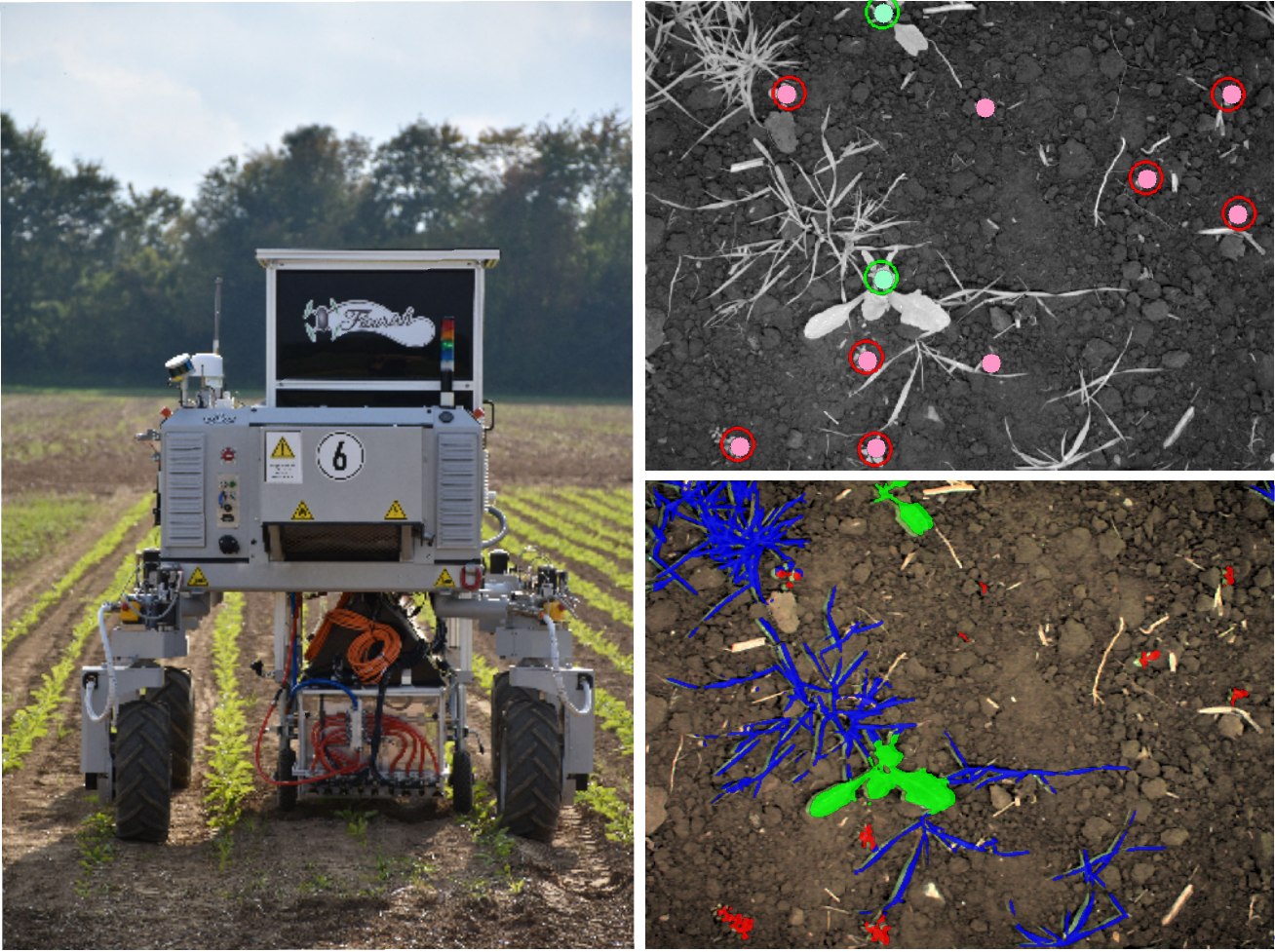}
 \caption{BoniRob platform operating in a sugar beet
  field while performing weed control through selective spraying and 
  mechanical weed stamping. Our approach analyzes camera images
  and provides two classification outputs: the stem positions of the
  crop plants (green) and \dicotyls (red) as well as a pixel-wise semantic
  segmentation of the input into the classes crop (green), \dicotyl (red),
  grass (blue) and soil (no color).}
  \label{fig:motivation}
  \vspace{-0.45cm}
\end{figure}

  
The main contribution of this paper is an end-to-end trainable pipeline for
joint pixel-wise plant segmentation and plant stem detection enabling plant-specific treatment, for example fertilizing a crop or destroying a weed. We
employ a fully convolutional neural network (FCN) architecture sharing the
encoded representation of the image content for the specific tasks, i.e., the
semantic segmentation of the crops, \dicotyls, and grasses, as well ass the
stem detection of the individual crops and \dicotyls for mechanical removal.
More specifically, we jointly estimate the pixel-wise segmentation into the
classes (1)~crop, (2)~\dicotyl, (3)~\grass, and (4)~background, i.e., mostly
soil, and estimate the stem locations of crops and \dicotyls at the same time.



In sum, we make the following two claims: Our approach is able to
(i)~determine the stem positions of crop and weed stems, and (ii)~accurately
separate \grass from \dicotyl for the purpose of a specific treatment in mind.
Furthermore, we show that our approach has a superior performance in
comparison to other state-of-the-art approaches, such as~\cite{haug2014ias,
lottes2016icra, milioto2018icra}. All claims are experimentally validated on
real-world data. Moreover, we plan to publish our code and the datasets used in
this paper.
  
\section{Related Work}
\label{sec:related}

In recent years, significant progress has been made towards vision based \cropvsweed 
classification systems, both using handcrafted features 
\cite{haug2014wacv}, \cite{lottes2016icra}, \cite{lottes2016jfr} and \mbox{end-to-end}
 methods based on convolutional neural networks (CNN) \cite{mccool2017icra}, \cite{sa2018ral},
\cite{milioto2018icra}. However, none of these methods estimate the
stem locations or other information which can be directly used for targeted
intervention. With our work, we aim to bridge this gap by developing a system which
integrates both the task of plant classification and stem detection in an 
\mbox{end-to-end} manner with the goal of targeted treatment in mind.

Other approaches have been developed to classify individual plants and 
identify their stem locations. Most of these approaches are developed 
based on manually designed heuristics with specific use cases in mind.
Kiani and Jafari~\cite{kiani2012jast} use hand-crafted shape features selected
on the basis of a discriminant analysis to differentiate corn plants from
weeds. They identify the stem position of the plant as the centroid of the
detected vegetation. This leads to sub-optimal results particularly when the
plant shapes are not symmetric or multiple plants are overlapping.
Midtiby \etal~\cite{midtiby2012be} present an approach for sugar beet by detecting
individual leaves and use the contours of the leaves for finding the stem
locations. This approach fails to locate the stems in the presence of
occluded leaves or overlapping plants.

Moving towards a machine learning based approach,
Haug~\etal~\cite{haug2014ias} propose a system to detect plant stems using
keypoint-based random forests. They employ a sliding window based
classifier to predict stem regions by using several hand-crafted geometric and
statistical features. Their evaluation shows that the approach often
misses several stems for overlapping plants or generates false positives for
leaf areas which locally appear to be stem regions. Kraemer \etal~
\cite{kraemer2017iros} aim at addressing this issue by increasing the field of
view of the classifier using a fully convolutional networks~(FCN)~\cite{long2015cvpr-fcnf}.
The goal of their work is to identify crop
stems over a temporal period allowing them to use the stem locations as
landmarks for localization. 

Our work overcomes many of the limitations by taking a holistic approach by
jointly detecting stems and estimating a pixel-wise segmentation of the plants
based on FCNs. Moreover, we explicitly distinguish crop and \dicotyl stems, since it 
enables plant-specific treatment, for example fertilizing a crop or destroying a weed mechanically.

\section{Joint Stem Detection and \Cropvsweed Semantic Segmentation}
\label{sec:main}

\newcommand{\encoder}{encoder\xspace}
\newcommand{\stemDecoder}{stem decoder\xspace}
\newcommand{\stemFeatures}{stem features\xspace}
\newcommand{\plantDecoder}{plant decoder\xspace}
\newcommand{\plantFeatures}{plant features\xspace}
\newcommand{\stemMask}{stem mask\xspace}
\newcommand{\plantMask}{plant mask\xspace}
\newcommand{\stemextraction}{stem extraction\xspace}

\begin{figure*}
  \centering
  \includegraphics[width=\linewidth]{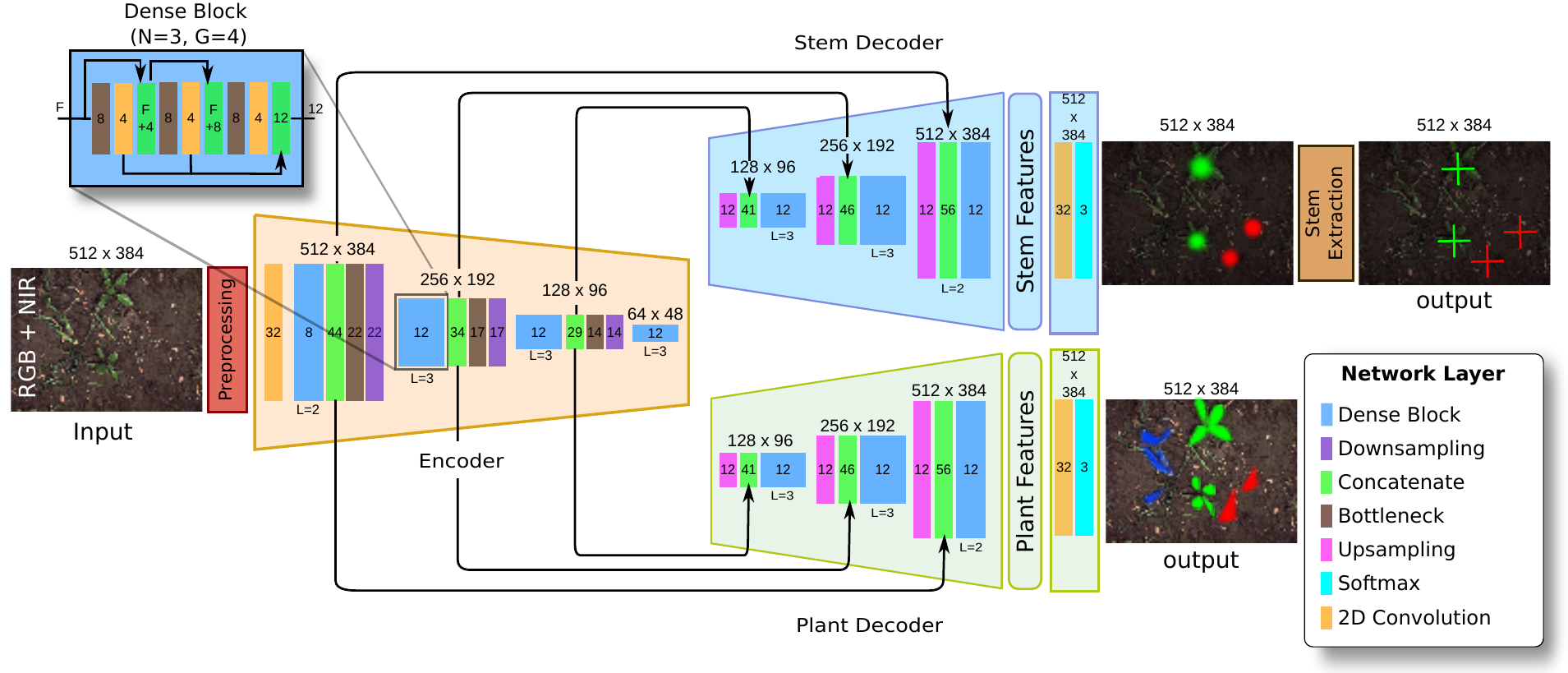}
  \caption{Architecture of our approach as described in \secref{sec:archconcept}.
  We first encode the input image using the \encoder. Then pass the encoded feature
  volume to the task-specific decoder, the \stemDecoder and the
  \plantDecoder. Thus, we obtain two outputs, the \plantMask for the
  semantic segmentation of the plants and the \stemMask for the
  segmentation crop-weed stem regions. Finally, we extract the stem positions from the \stemMask 
  in the \stemextraction. Note that we denote the size of the feature map above each block of layers. 
  Inside the layers, we show the number of output features maps.}
\label{fig:archconcept}
\end{figure*}  

The main objective of our plant classification system is to simultaneously
provide a semantic segmentation of the visual input into the classes crop,
\dicotyl, \grass, and soil as well as the stems positions for 
\dicotyls  and crops. The stem positions are a prerequisite in selective, high precision
treatments, e.g., by mechanical stamping or by laser-based weeding. 
The provided pixel-wise label mask provides the area for more
granulated approaches such as selective spraying. We propose an approach for
the joint processing of the plant classification and the stem detection task
based on FCN. Our network architecture
shares the encoded features for classifying the
stem regions as well as for the pixel-wise semantic segmentation 
using two, task-specific decoder networks.

The input to our network is  either given by the raw RGB or by RGB plus a
 near infra-red~(NIR) channel. The output of the proposed network
consists of two different label masks representing a probability distribution
over the respective class labels. The first output is the \plantMask
reflecting the pixel-wise semantic segmentation of the vegetation in image 
space, whereas the second output is the \stemMask segmenting regions within
the image, which correspond to crop stems and weed stems.  Finally, we
extract pixel-accurate stem positions from the \stemMask.

\subsection{General Architectural Concept}
\label{sec:archconcept}

\figref{fig:archconcept} depicts the proposed architecture of our joint plant and
stem detection approach. The main processing steps of our approach are the
(i)~preprocessing (red), the (ii)~\encoder (orange), the (iii)~\plantDecoder
(blue), the (iv)~\stemDecoder (green), and (v)~the \stemextraction (brown).

As common practice in semantic segmentation \cite{badrinarayanan2017pami,
jegou2017arxiv}, the \encoder uses convolutional layers
and downsampling operations to extract a compressed, but highly informative,
representation of the image. We use this encoded image representation as input
to our task-specific decoders, i.e., a \plantDecoder and a \stemDecoder.
The \plantDecoder produces the \plantFeatures, for the
segmentation of the soil, crops, \dicotyls and \grasses, whereas the
\stemDecoder produces the \stemFeatures for the segmentation of the \cropvsweed 
stem regions. Both decoders upsample the shared code volume back to the
original input resolution to allow for a pixel-wise segmentation. Finally, we
further analyze the \stemMask containing the segmentation of potential stem
region  and extract the stem locations of crops and \dicotyls from it.
In the following sections, we describe the different parts of the proposed
pipeline in more detail.

\subsection{Preprocessing}
\label{sec:preprocessing}

To improve the generalization capabilities and also aid the convergence of training,
we first preprocess each channel of the given input images, i.e.,
red, green, blue, and near infra-red, as follows. First, we apply a Gaussian
smoothing using a $[5\times5]$ kernel with weights from a Gaussian with mean
$\mu = 0$ and $\sigma^2 = 1$. Then, we standardize each channel by substracting  
the mean and divide it by its variance. Finally, we contrast
stretch the input values to the range $[-0.5,0.5]$.

\subsection{FC-DenseNet}

The main building block of our FCN architecture is inspired by the so-called
Fully Convolutional DenseNet (FC-DenseNet)~\cite{jegou2017arxiv}, which
combines the recently proposed densely connected CNNs organized as dense
blocks~\cite {huang2017cvpr-dccn} with fully convolutional networks (FCN)~\cite {long2015cvpr-fcnf}.

A dense block is given by a stack of $N$~subsequent convolutional layers
operating on feature maps with the same spatial resolution. Here, we define a
convolutional layer as composition of a $[3 \times 3]$ convolution, leaky
rectified linear units~\cite{Maas2013icml-ws}, batch normalization~\cite{szegedy2016cvpr}, 
and dropout~\cite{srivastava2014jmlr}. 
Each convolutional layer gets as input a concatenation of the result of the previous 
layers. For computational efficiency, we use a bottleneck layer before
a convolutional layer implemented by a $[1\times 1]$ convolution ~\cite{lin2014iclr}. 
A dense block produces $N\cdot G$ feature maps, where $G$
is the growth rate~\cite {huang2017cvpr-dccn}.

We use subsequent dense blocks inside the encoder and concatenate the
input to a dense block with its  output, which subsequently is  compressed again
by a bottleneck layer to reduce the growth of feature maps within the encoder.
Each dense block is followed by a downsampling operation using strided
convolutions employing a convolutional layer with a $[5\times5]$ kernel and a
stride of $2$.

From the encoded and compressed information, we generate two separate
feature volumes specialized for pixel-wise plant classification and stem
detection. Thus, we have two decoders, which perform an upsampling using a
strided transpose convolution \cite{dumoulin2016arxiv} with $[2\times2]$
kernel and a stride of $2$. Both decoders also use dense blocks as their main
building block and follow the same architectural design to produce the
\plantFeatures and \stemFeatures.
Moreover, both task specific decoders use feature maps produced by the encoder
through skip connections. We concatenate the corresponding feature maps
sharing the same spatial resolution from the encoder before we again use dense
blocks for feature computation. Skip connections from the encoder to the
decoders facilitates the recovery of spatial information~\cite{badrinarayanan2017pami}. Finally, we transform the feature maps produced by the \stemDecoder and the
\plantDecoder into the pixel-wise probability distribution over their
respective class labels by a $[1\times1]$ convolution followed by a softmax layer.
\pagebreak

For learning, we use a multi-task loss $L$ combining the loss for the
plant segmentation $L_{\text{plant}}$ and for the stem region segmentation
$L_{\text{stem}}$, i.e.,
\begin{align}
 L &= (1-\alpha)\cdot L_{\text{stem}} + \alpha\cdot L_{\text{plant}},
\end{align} 
where we use $\alpha = 0.5$. $L_{\text{plant}}$ is a weighted cross entropy
loss, where we penalize errors regarding the crops, \dicotyls and grasses by a
factor of 10. $L_{\text{stem}}$ is a loss based on an approximation of the
intersection over union~(IoU) metric as it is more stable with imbalanced
class labels \cite{Rahman2016icsv}, which is the case in our problem with
under-represented stems as compared to the amount of soil. The multi-task loss
also enables to share information for learning the encoder, which can use the
loss information from both decoders in the backward pass of the
backpropagation.
 
\subsection{Stem Extraction}

Given the pixel-wise stem mask prediction from the neural network, i.e., $P(y|\vec{x})$ with $y
\in \{\text{soil},\text{crop}, \text{\dicotyl}\}$ for each pixel $\vec{x}$, we want to
extract a stem location for the crops and the \dicotyls. To
this end, we first determine for each pixel the class with highest label
probability, i.e.,~\mbox{$y^* = \argmax_y P(y|x)$}. Next, we determine the
connected components $\set{X}^{c}_j$ for each class $c$ and compute the weighted mean $\bar{\vec{x}}^c_j$ of the pixel locations by
\begin{align}
 \bar{\vec{x}}^c_j &= \frac{\sum_{\vec{x} \in \set{X}^c_j} P(y = c|\vec{x}) \cdot \vec{x}}{\sum_{\vec{x} \in \set{X}^c_j} P(y = c|\vec{x})}.
\end{align}
The weighted means $\bar{\vec{x}}^c_j$ for class $c$ are then the stem detections reported by our approach. 

\pagebreak

\section{Experimental Evaluation}
\label{sec:exp}

\begin{figure}
\setlength\tabcolsep{4pt}
  \begin{tabular}{ccc}

      \multicolumn{2}{c}{\small\dataA} & \small{\dataB} \\
      {\includegraphics[width=0.31\linewidth]{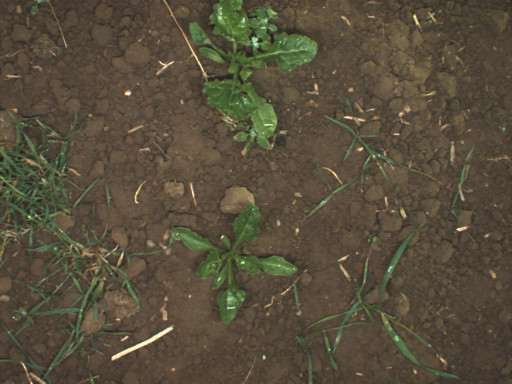}}
  \hspace{-3.5mm}
   &
      {\includegraphics[width=0.31\linewidth]{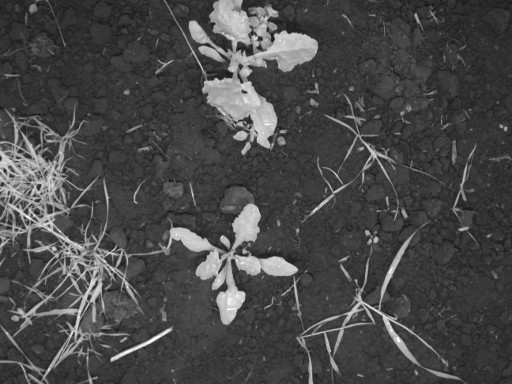}}
  \hspace{-2mm}
   &
      {\includegraphics[width=0.31\linewidth]{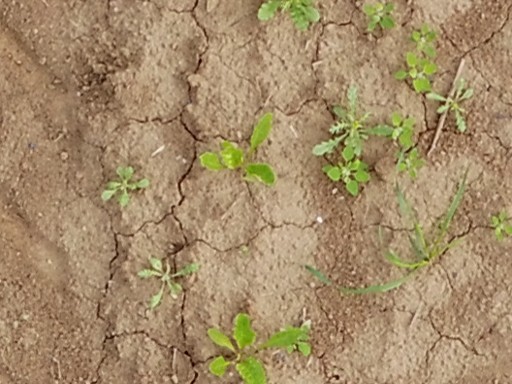}}\\

  \end{tabular}
\caption{Examples images. The left RGB+NIR image shows data from the \dataA. The right RGB image belongs to the
\dataB. Note, the difference in the lighting, but also soil conditions.}
\label{fig:dataset}
\vspace{-0.5cm}
\end{figure}

%
Our experiments are designed to show the capabilities of our method and to
support our two claims: (i)~Our approach is able to detect accurately the
stem locations of crops and \dicotyls, and simultaneously (ii)~is able to
accurately segment the images into the classes crops, \dicotyl, grasses and
soil.

The experiments are conducted on data from two different sugar beet fields located near
Bonn. Both datasets contain sugar beet plants, which are our crop, different 
\dicotyls and \grasses. The first dataset, called \dataA, consists of $382$  RGB+NIR images
recorded under artificial lighting conditions with the BOSCH DeepField
Robotics BoniRob and is a subset of a publicly available dataset~\cite{chebrolu2017ijrr}. It contains $488$ crop stems and $1,607$ weed stems.
The second dataset, called \dataB, contains $400$ RGB images recorded with an unmanned aerial vehicle~(UAV), 
the DJI~Inspire~II. In sum, it contains $808$ crop stems and
$1,451$ \dicotyl stems. Both datasets represent challenging conditions for
our approach as they
contain several different \dicotyl types of different sizes and multiple grass types and
have a substantial amount of inter-plant overlap. \figref{fig:dataset}
shows example images of each dataset.

We use the mean average precision~(mAP) over the per-class average precisions
(AP) \cite{everingham2010ijcv} as metric for our evaluation. The mAP represents
the area under the interpolated precision-recall curve. As noted by
Everingham \etal~\cite{everingham2010ijcv}, a method must have a high precision at all levels of recall
to achieve a high score with this metric. For the stem detection task, a
predicted stem is considered to be a positive detection if its Euclidean
distance to the nearest unassigned ground truth stem is below a threshold
$\theta = 10$\,mm corresponding to the size of the mechanical stamping tool of
the BoniRob. Furthermore, we compute the mean average distance~(MAD) for all
true positives to show the spatial precision of our approach. For the
segmentation task, we evaluate the performance in a pixel-wise manner also using the mAP.

For comparison, we also evaluate the performance with respect to other methods. For the
stem detection, we refer to the \emph{baseline-stem} approach, where we apply
our proposed architecture as a single encoder-decoder FCN. Analogously, we refer
to the \emph{baseline-seg} approach, when using our architecture only for the
semantic segmentation task. We furthermore compare the performance with our
implementations of state-of-the-art approaches for stem detection and \cropvsweed 
classification. For stem detection, we re-implemented the approach of Haug
\etal \cite{haug2014wacv} using a random forest and the described features,
which we denote by RF. Next, we use the same methodology as Haug \etal, but
with the visual and shape features proposed by Lottes \etal
\cite{lottes2016icra} and denote this method by RF+F. 
For the segmentation task, we use the approach by Lottes \etal
\cite{lottes2016icra} and a state-of-the-art FCN-based approach
\cite{milioto2018icra} for \cropvsweed detection as a reference.

\subsection{Parameters}

In all our experiments, we learned our network from scratch using only the
training data of the respective dataset, which is $75$\,\% of all available
images. For validation, we use $5$\,\% and a $20$\,\%-held out portion for
evaluating the test performance, which we report within this section. We
downsample the images to a resolution of $W=512$ and $H=384$ pixels, which
yields a ground resolution of approx. $1\,\frac{\text{mm}}{\text{pixel}}$. In
all experiments, we use a growth rate $G=4$ and dropout probability
$p=\frac{1}{3}$. Following common best practices for training deep networks,
we initialize the weights according to He \etal~\cite{he2015iccv}, use ADAM
for optimization, and a mini-batch size of $B=4$. The initial learning rate is
set to $0.01$ and divided by $10$ after $50$, $250$ and $1,000$ epochs. We
stop training after $2,000$ epochs. We implemented our approach using Keras.

\subsection{Stem Detection Performance}

\small
\tabcolsep=0.11cm
\begin{table}[t]
\caption{Stem Detection Performance on \dataA}
\centering
\begin{tabular}{cC{1cm}|C{1cm}C{1cm}|C{1cm}C{1cm}}
\toprule
\multirow{2}{*}{Approach} & &  \multicolumn{2}{c|}{Crop} & \multicolumn{2}{c}{Dicot}\\ 

        & mAP & AP  & MAD & AP & MAD \\
  \midrule
      ours                       & 79.2  & 93.5 & 2.5 & 65.0 & 1.8 \\
      RF+F \cite{lottes2016icra} & 51.1  & 85.1 & 1.3 & 17.1 & 1.7 \\
      RF \cite{haug2014wacv}     & 48.8  & 67.5 & 2.2 & 28.9 & 2.1\\
      baseline-stem             & 73.8  & 88.9 & 2.7 & 58.8 & 1.9\\
      \bottomrule

\end{tabular}
\label{tab:stem_datasetA}
  \vspace{-2mm}

\end{table}

\small
\tabcolsep=0.11cm
\begin{table}[t]
\caption{Stem Detection Performance on \dataB}
\centering
\begin{tabular}{cC{1cm}|C{1cm}C{1cm}|C{1cm}C{1cm}}
\toprule
\multirow{2}{*}{Approach} & &  \multicolumn{2}{c|}{Crop} & \multicolumn{2}{c}{Dicot}\\ 

        & mAP & AP  & MAD & AP & MAD \\
  \midrule
      ours                       & 75.3 & 78.8 & 3.2 & 71.5 & 2.1 \\
      RF+F \cite{lottes2016icra} & 51.8 & 60.6 & 3.1 & 70.2 & 1.8\\
      RF \cite{haug2014ias}     & 49.1 & 59.5 & 2.7 & 38.7 & 1.7\\
      baseline-stem             & 75.7 & 78.8 & 3.2 & 72.6 & 2.1\\
      \bottomrule

\end{tabular}
\label{tab:stem_datasetB}
  \vspace{-2mm}

\end{table}
\normalsize

 \begin{figure*}
 \centering
 \tabcolsep=0.5mm
  \begin{tabular}{ccccc}
  
      \small{Image} & \small{Ground Truth} & \small{Ours} &  \small{RF \cite{haug2014ias}} &  RF+F \cite{lottes2016icra} \\
      
      {\includegraphics[width=0.194\linewidth]{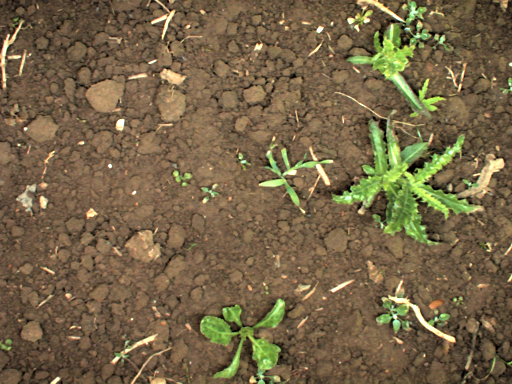}}
   &
      {\includegraphics[width=0.194\linewidth]{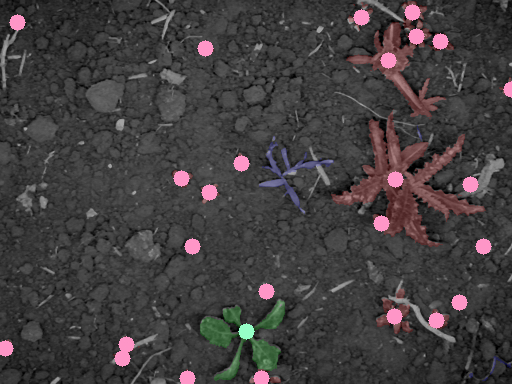}}
   &
      {\includegraphics[width=0.194\linewidth]{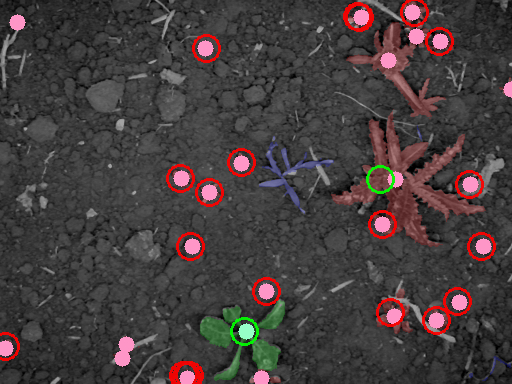}}
   &
      {\includegraphics[width=0.194\linewidth]{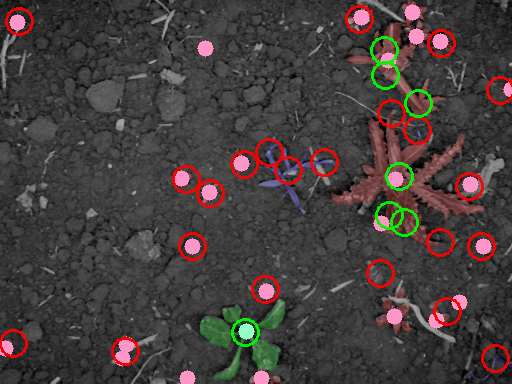}}
   &
      {\includegraphics[width=0.194\linewidth]{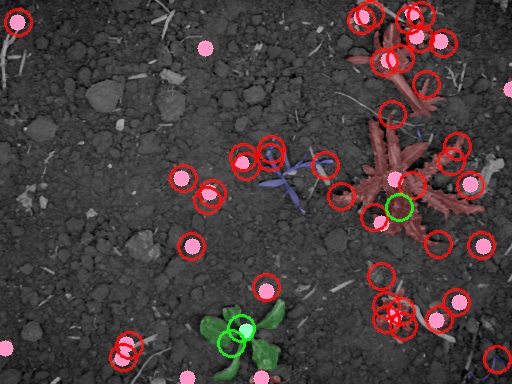}}
\\
      {\includegraphics[width=0.194\linewidth]{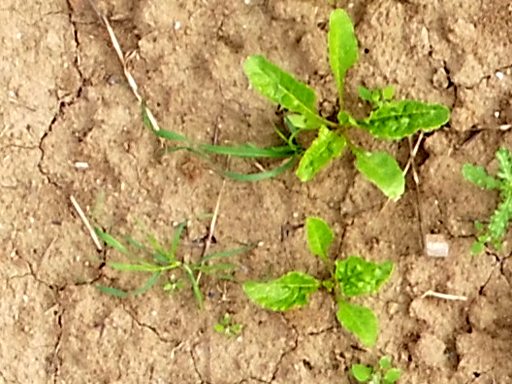}}
   &
      {\includegraphics[width=0.194\linewidth]{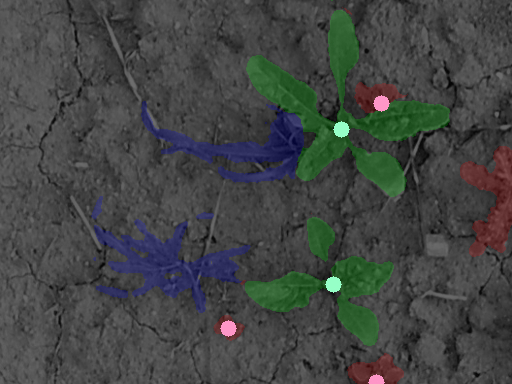}}
   &
      {\includegraphics[width=0.194\linewidth]{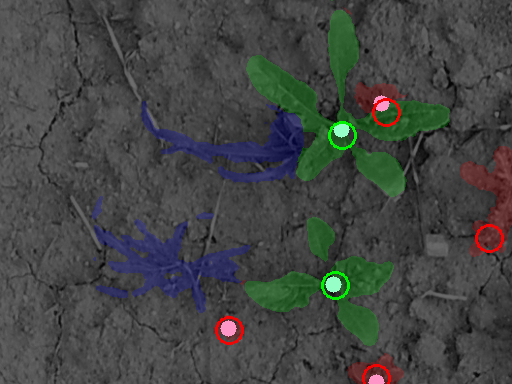}}
   &
      {\includegraphics[width=0.194\linewidth]{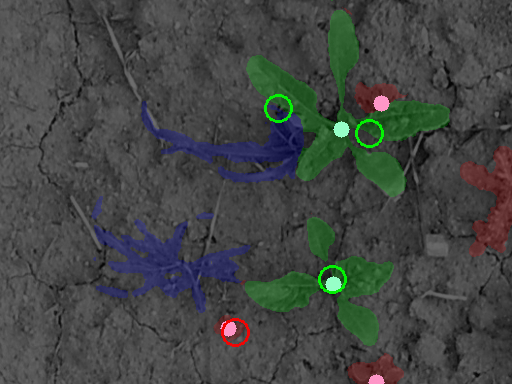}}
   &
      {\includegraphics[width=0.194\linewidth]{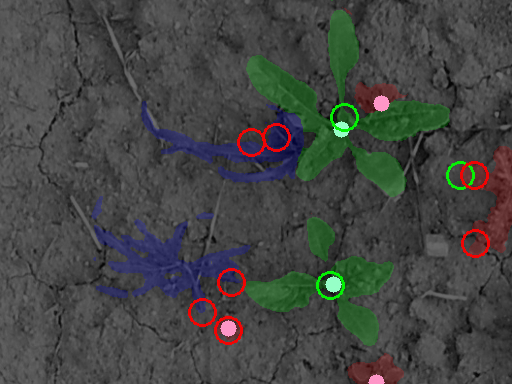}}

  \end{tabular}
 \caption{Qualitative results of the stem detection for \dataA (top row) and
 \dataB (bottom row). From left to right: RGB input image, ground truth
 information for the pixel-wise class labels provided for better view,
 predicted stems and corresponding ground truth by our approach, RF
 \cite{haug2014ias} and RF+F \cite{lottes2016icra}. Red refers to the class
 \dicotyl and green refers to crop. the solid circles in the same, but brighter color, refer to the respective
 ground truth information.}
  \vspace{-2mm}
 \label{fig:qualStem}
\end{figure*}

 \begin{figure*}
  \centering
  \tabcolsep=0.5mm
  \begin{tabular}{ccccc}
      \small{Image} & \small{Ground Truth} & \small{Ours} &  \small{FCN+PF \cite{milioto2018icra}} &  RF+F \cite{lottes2016icra} \\
      
      {\includegraphics[width=0.194\linewidth]{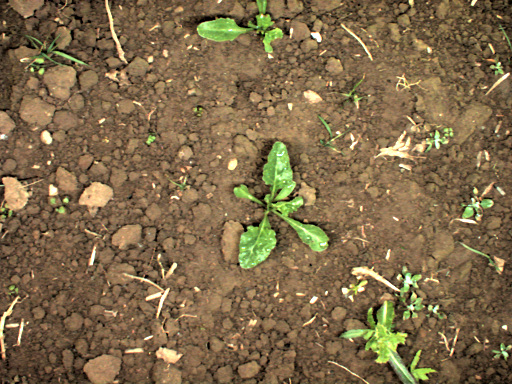}}
   &
      {\includegraphics[width=0.194\linewidth]{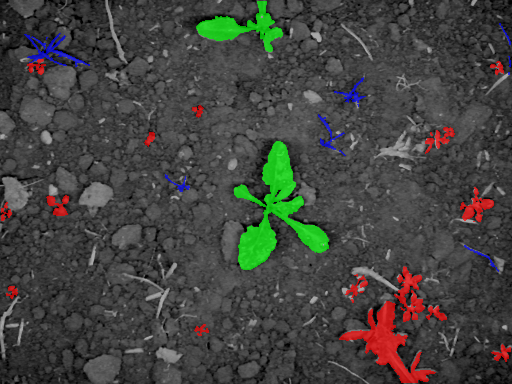}}
   &
      {\includegraphics[width=0.194\linewidth]{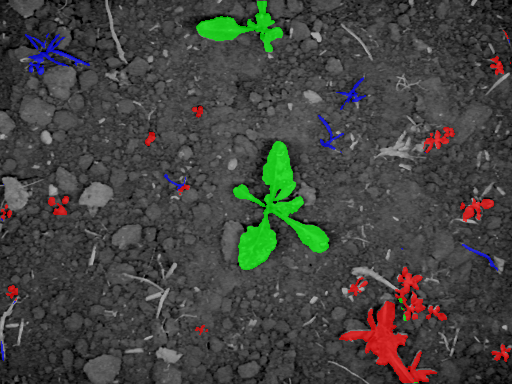}}
   &
      {\includegraphics[width=0.194\linewidth]{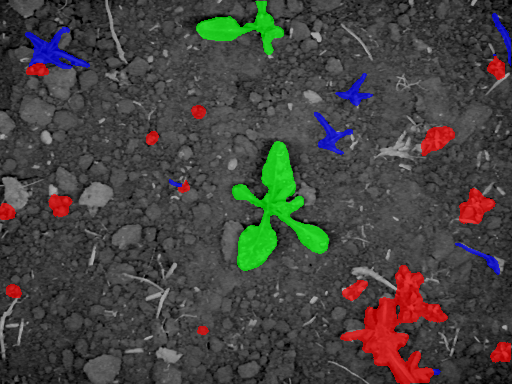}}
   &
      {\includegraphics[width=0.194\linewidth]{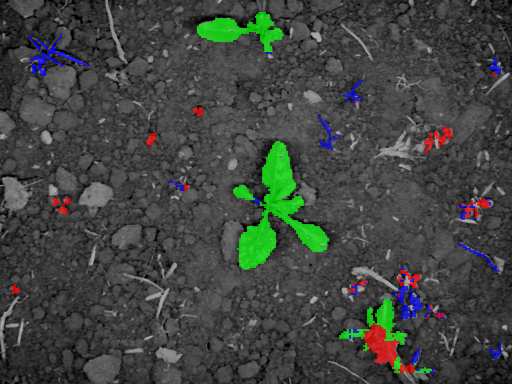}}
\\
      {\includegraphics[width=0.194\linewidth]{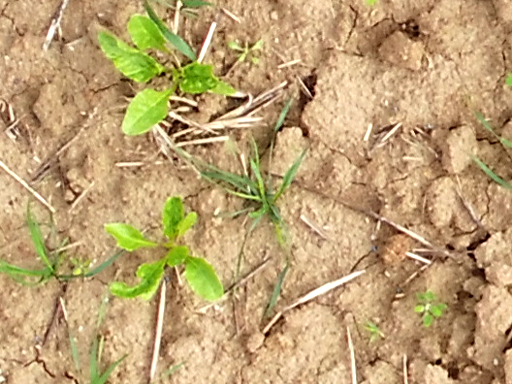}}
   &
      {\includegraphics[width=0.194\linewidth]{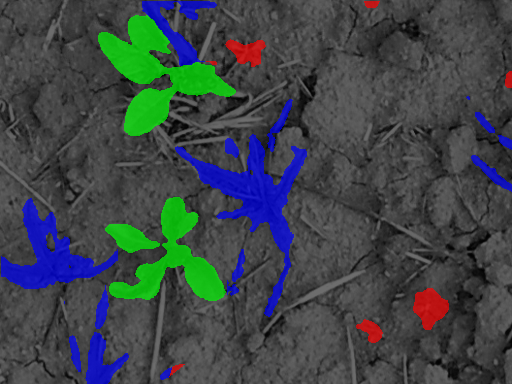}}
   &
      {\includegraphics[width=0.194\linewidth]{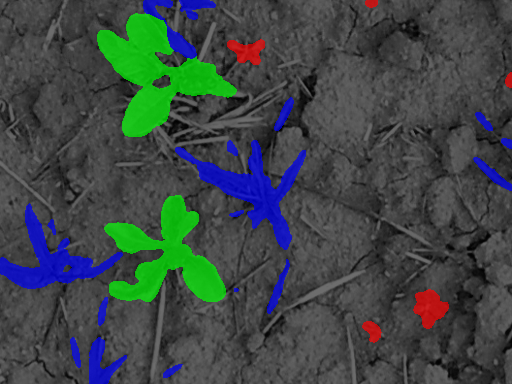}}
   &
      {\includegraphics[width=0.194\linewidth]{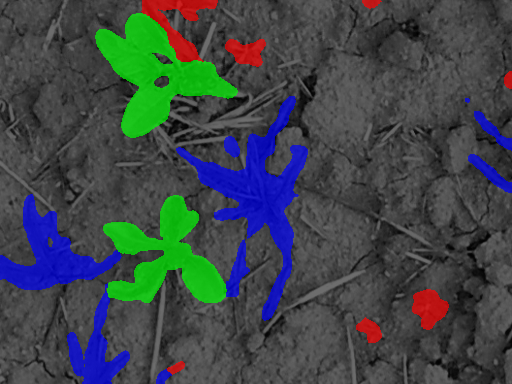}}
   &
      {\includegraphics[width=0.194\linewidth]{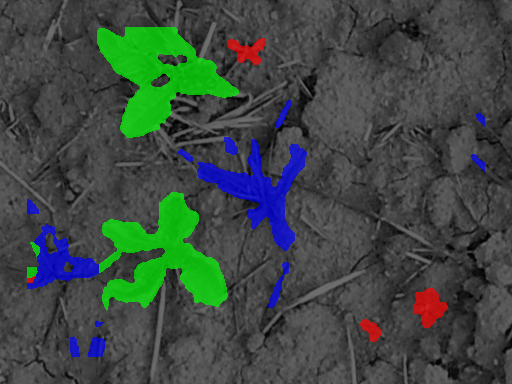}}

  \end{tabular}

 \caption{Qualitative results of the pixel-wise semantic segmentation for 
 \dataA (top row) and \dataB (bottom row). From left to right: RGB input
 image, the corresponding ground truth information for the pixel-wise class
 labels, prediction obtained by our proposed approach, FCN+PF
 \cite{milioto2018icra}, and RF+F \cite{lottes2016icra}. Red refers to the
 class \dicotyl, green refers to crop and blue refers to grass.}
\vspace{-2mm}
 \label{fig:qualSeg}
\end{figure*}

\small
\tabcolsep=0.11cm
\begin{table}[t]
\caption{Segmentation Performance on \dataA}
\centering
\begin{tabular}{cC{1cm}|C{1cm}C{1cm}C{1cm}C{1cm}}
\toprule
\multirow{2}{*}{Approach} & & {Soil} & {Crop} & Dicot & Grass \\ 

        & mAP & AP & AP & AP & AP \\
  \midrule
      ours                          & 83.8 & 99.8 & 91.2 & 69.4 & 75.0\\
      RF+F \cite{lottes2016icra}    & 68.9 & 97.7 & 85.1 & 46.4 & 46.4\\
      FCN+PF \cite{milioto2018icra} & 64.5 & 99.8 & 85.3 & 38.1 & 34.8\\
      baseline-seg                  & 81.9 & 99.8 & 94.2 & 60.4 & 73.0\\
      \bottomrule
\end{tabular}
\label{tab:seg_datasetA}
\vspace{-0.2cm}
\end{table}
\normalsize

\small
\tabcolsep=0.11cm
\begin{table}[t]
\caption{Segmentation Performance on \dataB}
\centering
\begin{tabular}{cC{1cm}|C{1cm}C{1cm}C{1cm}C{1cm}}
\toprule
\multirow{2}{*}{Approach} & & {Soil} & {Crop} & Dicot & Grass \\ 

        & mAP & AP & AP & AP & AP \\
  \midrule
       ours                          & 87.3 & 99.3 & 79.8 & 87.9 & 82.0\\ 
       RF+F \cite{lottes2016icra}    & 80.2 & 93.0 & 63.2 & 89.2 & 75.3\\
       FCN+PF \cite{milioto2018icra} & 85.5 & 99.9 & 79.8 & 86.7 & 75.6\\
       baseline-seg                  & 87.8 & 99.4 & 75.7 & 88.9 & 75.7\\
      \bottomrule
\end{tabular}
\label{tab:seg_datasetB}
\vspace{-0.4cm}

\end{table}
\normalsize

The first set of experiments is designed to support our first claim that our
approach detects the stem position for crops and \dicotyls. We
compare the performance with the aforementioned approaches based on random
forests (RF and RF+F) as well as with our baseline approach. \tabref{tab:stem_datasetA}
and \tabref{tab:stem_datasetB} shows the respective performance for \dataA and
\dataB.

In both datasets, we see that our proposed approach outperforms the
competing approaches using random forests classifiers. The difference in
mAP is mainly due to the improved performance with respect to the
\dicotyl stem detection. In both datasets, we observe a gain of around
$25-30$\% in the mAP. 
We also gain around $5$\% in mAP with respect to the baseline-stem approach on \dataA.
On \dataB the performance is comparable.
This suggests that the stem detection benefits from using the shared encoder
influenced by both the stem detection loss $L_{\text{stem}}$ and the plant
segmentation loss $L_{\text{plant}}$. We conclude that employing the joint
encoder aids the performance for the stem detection task. Furthermore, it is
computationally more efficient compared to using two separate networks as it
saves around $\frac{1}{4}$ of the parameters by sharing the encoder.

\figref{fig:qualStem} illustrates qualitative results of the stem
detection in comparison to the other approaches. We can see
that our approach performs best regarding the stem detection for the
\dicotyl. The random forest based approaches tend to detect more false
positives for crops and \dicotyl stems on the image parts containing
vegetation. The FCN-based approach most probably benefits from the learned
features providing a richer representation for the given task.

Notably, we had to manually fine-tune the vegetation detection for the random
forest-based approaches (RF and RF+F), since the automated thresholding for
the vegetation detection step did not lead to satisfactory results. This holds
especially for the \dataB as it does not provide the additional NIR
information, which typically aids the vegetation segmentation. In
contrast for our approach, we selected only one set of hyper-parameters, such
as the training schedule and initialization scheme, for training both
datasets.

In terms of the MAD, we see that all approaches provide the stem position
within around $2$\,mm in object-space, which is a sufficient accuracy for
precise, plant-specific treatment, like mechanical stamping.

\subsection{Semantic Segmentation Performance}

The second experiment is designed to show the performance of the
pixel-wise semantic segmentation and to support our second claim that our
approach provides an accurate segmentation of the images into the classes
crop, \dicotyl, grass, and background. Here, we compare again
with RF+F \cite{lottes2016icra}, but now let the random forest predict the
pixel-wise classification of the image. In addition to that we compare the
performance with a state-of-the-art approach \cite{milioto2018icra} 
employing FCNs, denoted by FCN+PF.

\tabref{tab:seg_datasetA} summarizes the performance obtained for \dataA.
Here, our approach achieves the best results. With a mAP of $83.8$\% most of
the plants are correctly segmented. Analogous to the stem detection
experiment, the better performance is mainly due to the high precision and 
recall for weed classes, i.e. \dicotyl and grass. 

\figref{fig:qualSeg} illustrates qualitative results of the semantic
segmentation for both datasets. The
analysis of the qualitative result shows that our approach properly segments
small weeds and grasses, whereas the RF+F~\cite{lottes2016icra} has visibly
more false detections and the FCN+PF~\cite{milioto2018icra}, tends to have 
more ``blobby'' prediction. In turn, this leads to a high recall for weeds, 
but decreases the precision for these classes. 

Regarding the comparison with the baseline-seg model, we observe a similar
behavior as for the stem detection, i.e., a better performance on \dataA and
comparable one for \dataB. These results show that our approach provides
state-of-the-art performance for the semantic segmentation task outperforming 
two separate FCNs.

\section{Conclusion}
\label{sec:conclusion}

In this paper, we presented a novel approach for joint stem detection and
\cropvsweed segmentation using a FCN. 
We see our approach as a building block enabling  farm robots to perform
selective and plant-specific weed treatment. Our proposed architecture enables
a sharing of feature computations in the encoder, while using two distinct
task-specific decoder networks for stem detection and pixel-wise semantic
segmentation of the input images. The experiments with two different datasets
demonstrates the improved performance of our approach in comparison to 
state-of-the-art approaches for stem detection and \cropvsweed classification.

\section*{Acknowledgments}
We thank R.~Pude and his team from the Campus Klein Altendorf for their great
support as well as A.~Kr\"au\ss ling, F.~Langer, and J.~Weyler
for labeling the datasets.


%


\bibliographystyle{plain}

\bibliography{glorified,new}

\end{document}